\documentclass[11pt]{article}
\usepackage{nodalida2023}
\usepackage{times}
\usepackage{url}
\usepackage{latexsym}

\usepackage[T2A,T1]{fontenc}
\usepackage[utf8]{inputenc}
\usepackage[russian,english]{babel}

\aclfinalcopy 

\usepackage{hyperref}
\usepackage{multirow}

\title{Sentiment Analysis Using Aligned Word Embeddings for Uralic Languages}

\author{Khalid Alnajjar \\
  Rootroo Ltd\\
  Helsinki, Finland \\
  {\tt name@rootroo.com} \\\And
  Mika Hämäläinen \\
  Rootroo Ltd\\
  Helsinki, Finland \\
  {\tt name@rootroo.com} \\\And
  Jack Rueter \\
  University of Helsinki \\
  Finland\\
  {\tt first.last@helsinki.fi} \\}

\date{}

\begin{document}
\maketitle
\begin{abstract}
 In this paper, we present an approach for translating word embeddings from a majority language into 4 minority languages: Erzya, Moksha, Udmurt and Komi-Zyrian. Furthermore, we align these word embeddings and present a novel neural network model that is trained on English data to conduct sentiment analysis and then applied on endangered language data through the aligned word embeddings. To test our model, we annotated a small sentiment analysis corpus for the 4 endangered languages and Finnish. Our method reached at least 56\% accuracy for each endangered language. The models and the sentiment corpus will be released together with this paper. Our research shows that state-of-the-art neural models can be used with endangered languages with the only requirement being a dictionary between the endangered language and a majority language.
\end{abstract}

\section{Introduction}

Most of the languages spoken in the world are endangered to one degree or another. The fact of being endangered sets some limitations on how modern NLP research can be done with such languages given that many endangered languages do not have vast textual resources available online, and even with the resources that are available, there is a question about the quality of the data resulting from a variety of factors such as fluency of the author, soundness of spelling and, on the lowest level, inconsistencies in character encoding (see \citealt{9617119d543e4742b68b948232a59d40}). 

This paper focuses on the following Uralic languages: Erzya (myv), Moksha (mdf), Komi-Zyrian (kpv) and Udmurt (udm). Unesco classifies these languages as definitely endangered \cite{moseley_2010}. In terms of NLP, these languages have FSTs \cite{rueter2020open,rueter2021overview}, Universal Dependencies Treebanks \cite{partanen2018first,rueter2018towards} (excluding Udmurt) and constraint grammars available in Giella repositories \cite{moshagen2014open}. For some of the languages, there have also been efforts in employing neural models in disambiguation \cite{ens-etal-2019-morphosyntactic} and morphological tasks \cite{hamalainen2021neural}. Out of these languages, only Erzya has several neural based models available such as machine translation models \cite{dale2022first}, a wav2vec model and a Stanza model \cite{qi2020stanza}.

In this paper, we present a method for translating word embeddings models from larger languages into the endangered languages in question. Furthermore, we fine-tune the models with language specific text data, align them and show results in a sentiment analysis task where no training data is provided in any of the endangered languages. We have made our data and models publicly available on Zenodo\footnote{https://zenodo.org/record/7866456}.

\section{Related work}

Apart from the work described earlier in the context of the endangered languages in question, there has been a lot of previous work on multilingual NLP where a model is trained in one language to sentence classification and then applied in the context of other languages. In this section, we will describe some of those approaches together with sentiment analysis approaches.

A recent paper demonstrates sentiment analysis on 100 languages \cite{yilmaz2021multi}. The authors use RoBERTa-XLM to extract feature vectors. These are then used in training a bi-directional LSTM based classifier model. Another line of work \cite{liu2015multi} compares several different multilabel classification methods on the task of sentiment analysis showing that RAkEL \cite{tsoumakas2010random} gave the best performance on raw token input. A recent paper \cite{hamalainen2022video} demonstrated promising results in French sentiment analysis on a model that was trained in English, Italian, Spanish and German. The approach relied on a multilingual BERT \cite{devlin-etal-2019-bert}. \citet{ohman2021validity} suggests that lexicon based approaches, while viable for endangered languages, are not particularly suitable for sentiment analysis.

In the context of cross-lingual NLP, there is work on POS tagging. For instance, \citealt{kim-etal-2017-cross} propose a new model that does not require parallel corpora or other resources. The model uses a common BLSTM for knowledge transfer and another BLSTM for language-specific representations. It is trained using language-adversarial training and bidirectional language modeling as auxiliary objectives to capture both language-general and language-specific information.

Another line of work by \citealt{xu2018unsupervised} focuses on cross-lingual transfer of word embeddings, which aims to create mappings between words in different languages by learning transformation functions over corresponding word embedding spaces. The proposed algorithm simultaneously optimizes transformation functions in both directions by using distributional matching and minimizing back-translation losses. This approach uses a neural network implementation to calculate the Sinkhorn distance, a distributional similarity measure, and optimize objectives through back-propagation.

For machine translation \citealt{chen-etal-2022-towards} demonstrate the importance of both multilingual pretraining and fine-tuning for effective cross-lingual transfer in zero-shot translation using a neural machine translation (NMT) model. The paper presents SixT+, a many-to-English NMT model that supports 100 source languages but is trained on a parallel dataset in only six languages. SixT+ initializes the decoder embedding and full encoder with XLM-R large \cite{conneau2020unsupervised} and trains encoder and decoder layers using a two-stage training strategy.

\section{Data}

We use two books, Suomi eilen ja nyt (\textit{Finland yesterday and now}) by \citet{sejtfin1997} and \foreignlanguage{russian}{Павлик Морозов} (Pavlik Morozov) by \citet{gubarev_1953} both of which are available in Finnish, Erzya, Moksha, Komi-Zyrian and Udmurt. The sentences of the books have been aligned across all the languages at the Research Unit for Volgaic Languages in University of Turku. 
The size of the corpus for each language can be seen in Table \ref{tab:corpussize}.

\begin{table}[!ht]
\centering
\footnotesize
\begin{tabular}{|l|l|l|}
\hline
            & tokens & sentences \\ \hline
Finnish     & 43k    & 3.1k      \\ \hline
Erzya       & 50k    & 3.6k      \\ \hline
Moksha      & 51k    & 3.4k      \\ \hline
Komi-Zyrian & 50k    & 3.3k      \\ \hline
Udmurt      & 53k    & 3.6k      \\ \hline
\end{tabular}
\caption{The corpus size for each language}
\label{tab:corpussize}
\end{table}

Out of the entire corpus, we annotate 35 negative sentences and 33 positive sentences for evaluation for Finnish. We use the alignment information to project this annotation for the rest of the languages as well and verify manually that the sentences express the same sentiment in each language. This forms our test corpus for sentiment analysis that consists of altogether 68 sentiment annotated sentences.

Furthermore, we lemmatize all the texts using the FSTs provided in UralicNLP \cite{hamalainen2019uralicnlp}. The corpus is lemmatized because we intend to translate and align a lemmatized word embeddings model. This also makes the overall approach more robust given that covering the entire morphology of a language would require us to have much larger corpora.

\section{Word embeddings}

Word embeddings capture the semantic and syntactic links between words by constructing vector representations of words. These vectors can be utilized to measure the semantic similarity between words, find analogous concepts, cluster words~\cite{hamalainen2019let,stekel2021word} and more. In this work, we use English and Finnish as the big languages that facilitate aligning and classifying words and sentences for the endangered languages. English has an overnumerous amount of linguistic resources, whether as raw text or labeled data, while the endangered resources that we are working with have translation dictionaries for Finnish. For this reason, we use Finnish as the intermediate language that bridges these endangered languages with English resources.

The English model that we utilize is trained on the English Wikipedia dump of February 2017 and Gigaword 5th edition\footnote{\url{http://vectors.nlpl.eu/repository/20/17.zip}}~\cite{fares-etal-2017-word}. For Finnish, we used recent word embeddings trained by \citet{finembeddings}. These embeddings have been trained on several Finnish newspapers. Both of these models have been trained on lemmatized text.

The English word vectors have a dimension size of 300, while the Finnish word vectors have a dimension size of 100. In order to make the dimension sizes of the two sets of embeddings compatible, dimensionality reduction is applied to the English embeddings using principal component analysis (PCA)~\cite{10.1162/089976699300016728}. This process reduces the dimensionality of the English embeddings to 100, allowing them to be compared and analyzed alongside the Finnish embeddings.

\subsection{Creation of embeddings}

We aim to create word embeddings for endangered languages, which currently lack pre-existing embeddings. We use dictionaries from GiellaLT\footnote{https://github.com/giellalt}, which we augment using graph-based methods to predict new translations through the Ve'rdd\footnote{\url{https://akusanat.com/verdd/}} platform~\cite{alnajjar2022using,alnajjar2021enhancing}. We present the number of dictionary translations from each endangered language to Finnish that we obtained from the base dictionaries and predictions in Table~\ref{tab:translations}.

\begin{table}[!ht]
\centering
\small
\begin{tabular}{|l|l|l|l|}
\hline
 & Translations & Predictions & Total \\
\hline
kpv & 10983 & 14421 & 25404 \\
\hline
mdf & 36235 & 3903 & 40138 \\
\hline
myv & 18056 & 5018 & 23074 \\
\hline
udm & 36502 & 6966 & 43468 \\
\hline
\end{tabular}
\caption{Number of translations and predictions from the source languages to Finnish}
\label{tab:translations}
\end{table}

To create embeddings for the endangered languages, we adopt a method of cloning the Finnish embeddings and substituting the Finnish lemma with its corresponding translation in the endangered language. Where translations were absent, we omitted the word vector. The resulting embeddings consist of 7,908, 10,338, 7,535, and 9,505 word vectors for kpv, mdf, myv, and udm, respectively. The lower number of word coverage can be attributed to multi-word expressions present in the dictionaries but not the embeddings.

In the next step of our study, we fine-tuned the word embeddings for both Finnish and the endangered languages by using two books as additional data sources. This involved expanding the vocabulary of each embeddings model whenever a new word was encountered in the data. We also adjusted the embeddings weights based on the co-occurrences of words in the text, using a window size of 5 and a minimum count of 5 for a word to be considered in the vocabulary. After completing this process, the vocabulary size of the endangered language embeddings were 10,396, 11,877, 9,030, and 11,080, in the same order as mentioned above.

\subsection{Alignment of embeddings}
Our goal here is to align the Finnish word embeddings with the English ones, followed by aligning the embeddings of endangered languages to the Finnish embeddings, in a supervised manner. This was achieved by creating alignment dictionaries and aligning the embedding spaces together similarly to \citet{alnajjar2021word}.

To align Finnish embeddings with English, we used the Fin-Eng dictionary by~\citet{ylonen2022wiktextract}, which is based on the March 2023 English Wiktionary dump. We also used the Finnish-English dictionaries provided by MUSE~\cite{conneau2017word}. Regarding the endangered languages, we use the XML dictionaries to align them with Finnish. We set aside 20\% of the Wiktionary and XML data for testing the alignments.

One thing that we have noticed is the lack of the words ``no'' and ``not'' in the English embeddings due to stopword removal. To address this, we appended a translation from ``not'' to ``nt'' in the Finnish-English alignment data used in the training stage. Whenever the text contained these words, they were automatically mapped to ``nt'' in the following steps of our research.

\begin{table*}[h!]
\centering
\footnotesize
\begin{tabular}{ccccccc}
\hline
\textbf{Language} & \textbf{Label} & \textbf{Precision} & \textbf{Recall} & \textbf{F1-Score}  & \textbf{Accuracy} \\ \hline
\multirow{2}{*}{eng} & neg & 0.77 & 0.76 & 0.76 & \multirow{2}{*}{0.76} \\
                     & pos & 0.75 & 0.76 & 0.76 & \\ \hline
\multirow{2}{*}{fin} & neg & 0.77 & 0.75 & 0.76 & \multirow{2}{*}{0.75} \\
                     & pos & 0.73 & 0.75 & 0.74 & \\ \hline
\multirow{2}{*}{kpv} & neg & 0.57 & 0.57 & 0.57 & \multirow{2}{*}{0.56} \\
                     & pos & 0.55 & 0.55 & 0.55 & \\ \hline
\multirow{2}{*}{mdf} & neg & 0.63 & 0.65 & 0.64 & \multirow{2}{*}{0.63} \\
                     & pos & 0.64 & 0.62 & 0.63 & \\ \hline
\multirow{2}{*}{myv} & neg & 0.71 & 0.69 & 0.70 & \multirow{2}{*}{0.69} \\
                     & pos & 0.67 & 0.69 & 0.68 & \\ \hline
\multirow{2}{*}{udm} & neg & 0.69 & 0.63 & 0.66 & \multirow{2}{*}{0.63} \\
                     & pos & 0.58 & 0.63 & 0.60 & \\ \hline
\end{tabular}
\caption{Precision, recall, f1-score and accuracy for each language and label}
\label{table:precision-recall-f1}
\end{table*}

We followed the approach described by MUSE~\cite{conneau2017word} to align all the embeddings, with 20 iterations of refinement to align Finnish with English and 5 iterations to align all the other languages to Finnish.

\section{Sentence embeddings}

Word embeddings represent the meaning of a single word, whereas sentence embeddings represent the meaning of an entire sentence or document. Sentence embeddings are capable of capturing more the context and excel at tasks that call for comprehension of the meaning of a whole text, such as sentiment analysis. Hence, we build sentence embeddings for English that are based on the English word embeddings.

The procedure for creating sentence embeddings was conducted by averaging the word embeddings of a given sentence and subsequently feeding them to two fully-connected feed-forward layers, thereby constructing a Deep Averaging Network (DAN). The sentence embeddings are trained on the STS Benchmark~\cite{cer-etal-2017-semeval} using SBERT, a method for sentence embeddings that was proposed by~\cite{reimers-2019-sentence-bert}.

\section{Sentiment analysis}
We create a sentiment classifier that takes in the sentence embeddings and predicts a sentiment polarity label. For training the sentiment analysis model, we use the Stanford Sentiment Treebank \cite{socher2013recursive}, Amazon Reviews Dataset \cite{mcauley2013hidden} and Yelp Dataset\footnote{\url{https://www.yelp.com/dataset}}. These datasets are available in English and we use their sentiment annotations (positive-negative) to train our model.

The sentiment classifier is constructed as a three-layer fully-connected network, wherein the hidden layers are comprised of 300 neurons each. In order to mitigate overfitting, a dropout operation~\cite{JMLR:v15:srivastava14a} is performed prior to the final classification layer. The model consists of 121,202 trainable parameters in total, and is trained over the course of three epochs.

\section{Results}

In this section, we show the results of the sentiment classification model on the in-domain, English-language train splits of the sentiment corpora we used to train the model. Furthermore, we show the results of the sentiment classification model when applied on our own annotated data for the 4 endangered Uralic languages in question and Finnish. These results can be seen in Table \ref{table:precision-recall-f1}.

All in all, our model performs relatively well. The accuracy for Finnish is almost as high as it is for English despite not having any Finnish sentiment annotated training data. This means that our approach can achieve rather good results when there is a lot of translation data available between the two languages. The results drop for the endangered languages, but we do find the 69\% accuracy for Erzya to be quite formidable, however, the result for Komi-Zyrian of 56\% leaves some room for improvement.

\section{Conclusions}

In this paper, we outlined a method for translating word embeddings from a majority language, Finnish, to four minority languages - Erzya, Moksha, Udmurt, and Komi-Zyrian. The word embeddings were aligned and a new neural network model was introduced. This model was trained using English data to carry out sentiment analysis and was then applied to data in the endangered languages using the aligned word embeddings. 

We built an aligned sentiment analysis corpus for the four endangered languages and Finnish and used it to test our model. The results were promising and our study demonstrated that even the latest neural models can be utilized with endangered languages if a dictionary between the endangered language and a larger language is available.

\section*{Acknowledgments}

This research is supported by FIN-CLARIN and Academy of Finland (grant 345610 \textit{Kielivarojen ja kieliteknologian tutkimusinfrastruktuuri}).

\bibliographystyle{acl_natbib}
\bibliography{nodalida2023}

\end{document}